\documentclass[final]{cvpr}

\usepackage[accsupp]{axessibility}  %

\usepackage{times}
\usepackage{epsfig}
\usepackage{graphicx}
\usepackage{amsmath}
\usepackage{amssymb}

\usepackage[pagebackref,breaklinks,colorlinks]{hyperref}

\def\cvprPaperID{****} %
\def\confName{ICCV workshop}
\def\confYear{2023}
\graphicspath{{./figs/}{./figs/result/}}

\usepackage{times}
\usepackage{epsfig}
\usepackage{lipsum}
\usepackage{framed,multirow}
\usepackage{booktabs}
\usepackage{url}
\usepackage{xcolor}

\usepackage{comment}

\usepackage{algorithm}
\usepackage[noend]{algorithmic}
\usepackage{eqparbox}

\usepackage{graphicx}
\usepackage{amsmath,amssymb} %
\usepackage{xfrac}  %
\usepackage{bm}
\usepackage{booktabs}
\usepackage{rotating,multirow}
\usepackage{wrapfig}

\usepackage{tikz}
\usepackage[export]{adjustbox}

\graphicspath{{./Images/}{./Figures/result/}{./Images/result/}}

\newcommand{\pif}[1]{%
\begin{tikzpicture}
\draw (0,0) circle (1ex);
\end{tikzpicture}%
}
\newcommand{\pii}[1]{%
\begin{tikzpicture}
\draw[densely dotted] (0,0) circle (1ex);\draw (1ex,0) arc (0:#1:1ex) -- (0,0) -- cycle;
\end{tikzpicture}%
}
\newcommand{\pie}[1]{%
\begin{tikzpicture}
\draw[densely dotted] (0,0) circle (1ex);\draw[fill] (1ex,0) arc (0:#1:1ex) -- (0,0) -- cycle;
\end{tikzpicture}%
}
\newcommand{\pha}[1]{\phantom{\pie{#1}}}

\newcommand{\hlefte}{%
\begin{tikzpicture}
\draw (0, 0) -- (0ex, 0ex) arc(90:270:1ex) -- (0, 0);
\end{tikzpicture}%
}
\newcommand{\hleft}{%
\begin{tikzpicture}
\draw[fill] (0, 0) -- (0ex, 0ex) arc(90:270:1ex) -- (0, 0);
\end{tikzpicture}%
}
\newcommand{\hleftk}{%
\begin{tikzpicture}
\draw[densely dotted] (0, -1ex) -- (-1ex, -1ex) arc(180:270:1ex) -- (0, -1ex) ;
\draw[fill]  (0, 0) arc(90:180:1ex) -- (0, -1ex) -- (0, 0);
\end{tikzpicture}%
}
\newcommand{\hright}{%
\begin{tikzpicture}
\draw[fill] (0, 0) -- (0ex, 0ex) arc(-90:90:1ex) -- (0, 0);
\end{tikzpicture}%
}
\newcommand{\hrightk}{%
\begin{tikzpicture}
\draw[densely dotted] (0, 0) arc(-90:90:1ex) -- (0, 0);
\draw[fill](0, 1ex) -- (1ex, 1ex) arc(0:90:1ex); 
\end{tikzpicture}%
}

\begin{document}

\title{Self-training and multi-task learning for data exploitation: an evaluation study}
\title{Self-Training and Multi-Task Learning for Limited Data: \\ Evaluation Study on Object Detection}

\author{Hoàng-Ân Lê \qquad Minh-Tan Pham\\
IRISA, Université Bretagne Sud, UMR 6074, 56000 Vannes, France\\
{\tt\small \{hoang-an.le,minh-tan.pham\}@irisa.fr}
}

\maketitle

\begin{abstract}

Self-training allows a network to learn from the predictions of a
more complicated model, thus often requires well-trained teacher models
and mixture of teacher-student data %
while multi-task learning jointly optimizes different targets
to learn salient interrelationship and
requires multi-task annotations for each training example.
These frameworks, despite being particularly data demanding
have potentials for data exploitation if such assumptions can be relaxed.
In this paper, we compare self-training object detection 
under the deficiency of teacher training data
where students are trained on unseen examples by the teacher,
and multi-task learning with partially annotated data,~\ie
single-task annotation per training example. Both scenarios have
their own limitation but potentially helpful with
limited annotated data.
Experimental results show the improvement of performance
when using a weak teacher with unseen data for training
a multi-task student.
Despite the limited setup we believe the experimental results show the potential
of multi-task knowledge distillation and self-training, which
could be beneficial for future study. Source code is at
\url{https://lhoangan.github.io/multas}.

\end{abstract}

\section{Introduction}
\label{sec:intro}

Besides the impressive capability in solving complicated problems,
deep learning is well-known for 
being computationally expensive and highly data-demanding. %
The former, due to the complex architectural model, limits the 
deployment on low-capacity edge devices while
the latter constrains its generalizability and robustness.

The data scarcity problem is mostly due to expensive annotating
efforts as raw unlabeled data are practically everywhere~\cite{Chen2013web,Li2007online}.
Thus amelioration is often studied under a different training paradigm such as
self-, weakly- or unsupervised learning. Self-training is a
weakly-supervised method based on knowledge distillation or teacher-student
models~\cite{Hinton2015}. The idea is to  train a network,
called \textit{student} using the combination of
the same labeled data, with which a (usually) more cumbersome \textit{teacher}
network is trained, and new unlabeled data whose pseudo-labels are provided by
the teacher's predictions.

Self-training typically assumes that the same training data of the teacher is
part of the student training set~\cite{Radosavovic2018DDistil,Zoph2020ST} and
a sufficiently-trained teacher model can perform generally well on \textit{unseen} data,
making fewer errors than correct predictions~\cite{Radosavovic2018DDistil}.
However, the teachers' training data are not always available but only the
teacher's pre-trained weights due to copyright issues or confidentiality
concerns~\cite{Lopes2017DF,Nayak2019ZSKD}. To put an emphasis on the limit of
training data, in this paper, we consider the deficiency scenario where the teacher
is trained on \textit{small} amount of data and that the student training data have
not been seen by the teacher,~\ie the teacher-student training sets are disjoint.

On the other hand, multi-task learning assumes that simultaneously optimizing
multiple related targets for the same input helps the network to extract common
features, learn salient interrelationships, and improve performance.
Such assumption is usually perceived as demanding
and put extra burdens to data preparation as it multiplies the required annotations 
by the number of tasks and thus multiplies the efforts to maintain training set with
consistent annotations for all tasks. In the context of limited data, we consider the
multitask partially annotated scenario~\cite{Li2022MTPSL},
where the annotation sets are disjoint,~\ie each image is annotated
for a single task and there are no images containing both-task annotations.
This setup is data efficient and would be an alternative method to ameliorate 
data scarcity if a network could exploit task interrelationships to improve performance
without requiring all-task annotations per data example.

The contributions of the paper are as follows.
(1) We extensively study the performance of standard object detection with
self-training under limited annotated data.
The situation is studied together with feature-imitation
knowledge distillation as a way to mitigate the lack thereof.
The student's performances with respect to deficient teacher's
training status are observed by gradually reducing the training size.
(2) We briefly examine a cross-task scenario between object detection
and semantic segmentation for knowledge distillation where the teacher
and student are trained for different but related tasks.
(3) We evaluate the multi-task framework under
partially annotated scenario for data exploitation and show the
usefulness of combining both frameworks.
Although the paper does not provide novel architectural contributions nor
improvements over state-of-the-art results we believe the extensive experiments
provide important insights in resolving the data scarcity problem.

\section{Related Work}
\label{sec:related_work}

\subsection{Knowledge distillation}

The idea of knowledge distillation (KD) %
is based on the separability of the training and inference process.
Consequently, a network can learn from the outputs of a
larger model and get improved without complicated modification in
deployment. Since the pioneering publication of
Hinton~\textit{et al.}\cite{Hinton2015}, it has attracted various studies
and more understanding was obtained:
``a good teacher is patient and consistent''~\cite{Beyer2022Patient},
an intermediate-sized teaching assistance network helps better when
the complexity gap is large between the teacher and
student~\cite{Mirzadeh2020TA}, and how knowledge distillation can be
applied in  a self-supervised context with contrastive loss~\cite{Xu2020SSKD},~\textit{etc}.

On the one hand, knowledge distillation has been studied to 
accommodate task-oriented information
such as object localization~\cite{Zheng2022LD}, 
background-region knowledge~\cite{Guo2021Defeat}, and
teacher-student agreement~\cite{Zhang2021PDF} for object detection,
object detection in remote sensing data~\cite{Le2023KDRS},
or multi-task depth-semantic segmentation knowledge distillation~\cite{Li2020KDMT},~\textit{etc}.
On the other hand, it is at the
core of the self-training paradigm which seeks to expand
networks' learning capacity using unlabeled data.
The idea is that a (student) network can be improved by
training with the predictions of a pre-trained larger
(teacher) network on unlabeled data points.
For object detection, Radosavovic~\textit{et al.} proposes
a data-distillation model~\cite{Radosavovic2018DDistil}
which feeds various transformations of an unlabeled input
image to a well-trained teacher and uses the prediction
ensemble to train a student network.
Similarly, Zoph~\textit{et al.}~\cite{Zoph2020ST} 
studies the interaction between training methods and data augmentation
and compares self-training against pre-training using unlabeled ImageNet 
images.
Diverging from the previous work, we explore the student network performance
with respect to deficient teacher training data and examine
the behavior of feature-imitation knowledge distillation in combination
with self-training under such condition.

\subsection{Multi-task learning}

The main target of multi-task learning is to
infer simultaneously various aspects of a single input image.
The general idea is that the features for predicting each aspect or \textit{task}
of the same image should be overlapping, and by optimizing them in the same
model, using techniques such as attention mechanisms~\cite{Liu2019mtan}
or gating strategies~\cite{Bruggemann2020},
the network can pickup the interrelationships that benefits and complements
one another~\cite{Lu2021taskology,Li2022Universal}. 
Different task combinations would require different approaches to bring out
the shared information the tasks~\cite{Bruggemann2021}, inspiring different
cross-task studies~\cite{Lu2021taskology,Zamir2020xtaskConsist}.

Attempts have been made to relax the requirement for joint annotations of
all tasks as they put extra burdens to data preparation. Semi-supervised
learning methods such as~\cite{Chen2020meanTeacher,Imran2020}
lessen the data dependency and allow learning from unlabelled data,
yet all-task annotations per training sample are still required.

Multi-task partially learning, where each input image is annotated with
\textit{only} one of the tasks has been studied by Li~\etal~\cite{Li2022MTPSL}
for spatially dense tasks such as semantic segmentation and depth prediction.
The dense annotations of one task are projected to a joint task-space and provide
supervised signals for training the other task. This approach, however, is not
immediately applicable for sparse-annotation task such as object detection in this
paper.

\section{Implementation}
\label{sec:method}

We follow the setup and architectures described by
Zhang~\textit{et al.}~\cite{Zhang2021PDF}
for knowledge distillation and perform experiments on
the ResNet family. We employ the ResNet50
(in place of the ResNet34) backbone with PAFPN~\cite{Liu2018PAFPN}
neck for teacher and ResNet18 backbone with FPN 
neck for student. The first max-pooling layer of ResNet is removed
as in ScratchDet~\cite{Zhu2019scratchdet} and the context enhancement module
is added as in ThunderNet~\cite{Qin2019thundernet}.
The parameter ratio between the teacher and
the student is 1.61.

For object detection, the Retina-style~\cite{Lin2017focal} prediction
head is employed following~\cite{Zhang2021PDF}
with number of convolution blocks
reduced from 4 to 2. The intermediate feature channels are set to 256.
The detection head includes 2 identical sub-networks (except for the last layer)
for localization and classification.

The multi-scaled FPN features are aggregated for semantic segmentation using
the module by Kirillov~\etal~\cite{Kirillov2019panopticFPN}. The features at each scale
are passed through a sequence of convolution and 2x-upsampling modules until being
one-fourth of the input size. The subsequent
feature maps are element-wise added and upsampled to the input size. The intermediate features
are all 128-channel. The regular cross-entropy with softmax loss is employed.

The supervised training of the teacher and student model is
performed by the regular object detection losses,
including the balanced L1 loss for bounding-box localization~\cite{Pang2019}
and the quality focal loss~\cite{Li2020gfocal} for classification.
The teacher is also trained with mutual guide matching~\cite{Zhang2020},
instead of the regular intersection-over-union thresholds for a boost of performance.
The student self-training is performed by \textit{soft} supervised loss with the target
given by a teacher network: bounding-box localization uses the same balanced L1 loss
while classification uses focal loss~\cite{Lin2017focal} multiplied by $5\times10^{3}$ to be
in the similar range with localization\footnote{The weight is hard-coded and is the same
for all self-training experiments based on the initial loss values of localization
and classification (\eg 1.25 and 0.0002-0.0003, respectively), without being
exhaustively tested.}

For feature-imitation knowledge distillation, the feature maps after the neck 
layers are used (\textit{cf.}~\cite{Zhang2021PDF}). The KD losses to be
studied include (1) the trivial Mean Square Error (MSE) between the teacher's 
features and the projected maps of the student's, (2) the PDF-Distil~\cite{Zhang2021PDF}
measuring teacher-student disagreement, and (3) DeFeat~\cite{Guo2021Defeat}
distilling foreground and background regions separately.

For optimization, each task branch is trained alternately every iteration:
(1) After a mini-batch of input images with single-task annotations is passed through
the network, the loss function(s) of the corresponding task is computed
and back-propagating gradients through the task branch and the shared encoder;
(2) a mini-batch of the other-task images is passed through immediately in the next iteration;
(3) only after mini-batches from both tasks have been fed in and gradients accumulated are the network
parameters updated. Each experiment is trained for 70 epochs with early stopping using the validation set.

\begin{figure}[t]
    \footnotesize
    \protect
    \def\svgwidth{\linewidth}
    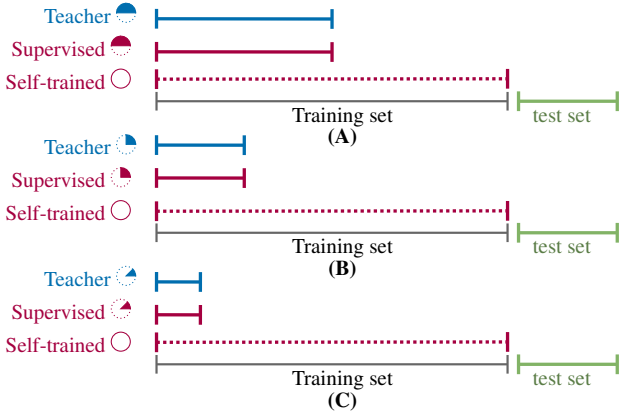
    \caption{Teacher-student data splits with overlapping data.
    Continuous lines indicate availability of annotated ground truths while the
    dotted lines indicate the lack thereof.
    }
    \label{fig:teacher_subset}
\end{figure}

\begin{figure*}[t]
    \centering
    \footnotesize
    \def\svgwidth{.90\linewidth}
    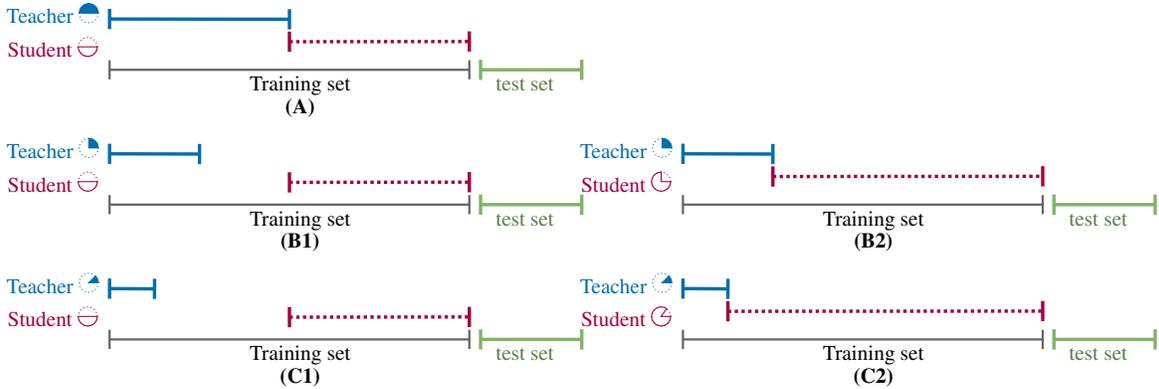
    \caption{Various scenarios of teacher-student disjoint subsets }
    \label{fig:disjoint_teacher_student}
\end{figure*}

\section{Experiments}
\label{sec:exp}

\subsection{Datasets}

Unless stated otherwise, we follow the conventional split of the Pascal VOC~\cite{PascalVOC} dataset
that keeps out the test split (4,952 images) of the VOC07 set for validation,
and uses the rest, training and validation split of both VOC07 and VOC12 set 
(16,551 images) for  training~\cite{Zhang2021PDF}.
We also show the results on the aerial vehicle detection VEDAI
dataset~\cite{vedai} for the first experiment.
The VOC evaluation metric for object detection is employed,
including AP50, AP75 and mAP~\cite{PascalVOC}.
For VEDAI, we report results on AP10 and AP25 due to the
very small-sized objects, following the suggestions
of~\cite{Le2022RS}. The IOU score~\cite{Jaccard1912}
is reported for the semantic segmentation task.
A detailed training split for a specific
experiment will be described at each experiment.

\subsection{Self-training for data exploitation}

In the first experiment, we confirm the usefulness of a teacher with unseen data
points, and consequently the idea of self-training. In particular,
we withhold the annotations of a half~\pii{-180} of the training set and train a supervised
\textit{teacher} network using the other half~\pie{180}. For the student, two cases are
compared%
, being trained (1) with the half annotated data~\pie{180} as the teacher's using the available
annotations (hard labels) and (2) with the entire training set~\pif{360} using only
teacher's prediction (soft labels) (Fig.~\ref{fig:teacher_subset}A).

The results in Table~\ref{tab:voc_vedai_half_vs_full} shows that using soft
label of the full dataset with possibly noisy soft labels is more beneficial
than using less data with ground truth targets.
PDF distillation shows superiority
for self-trained methods while DeFeat distillation is better for supervised methods.

The following subsections show ablation studies with the reduced sizes of the
available training set, weakening teachers performance and subsequently the
self-trained students.

\begin{table}[t]
    \centering
    \setlength{\tabcolsep}{2.8pt}
    \setlength\extrarowheight{.3ex}
    \begin{tabular}{@{}lcccc@{}}
         \toprule
         &&                             \multicolumn{3}{c}{VOC} \\
         \cmidrule(r){3-5}
         Model                 &           &      AP50   &      AP75     &     mAP   \\
         \midrule
         Teacher               & \pie{180} &     80.18   &     63.58     &     58.24 \\
         \midrule
         \midrule
         Supervised            & \pie{180} &     77.14   &     54.23     &     51.08 \\
         \phantom{Hard}+MSE    &           &     78.42   &     57.18     &     53.40 \\
         \phantom{Hard}+PDF    &           & \bf 78.97   &     57.67     &     54.18 \\
         \phantom{Hard}+DeFeat &           &     78.86   & \bf 58.15     & \bf 54.30 \\
         \midrule
         Self-trained          & \pif{360} &     79.95   &     60.06     &     55.64 \\
         \phantom{Soft}+MSE    &           &     79.51   &     60.61     &     55.53 \\
         \phantom{Soft}+PDF    &           & \bf 80.10   & \bf 61.22     & \bf 56.47 \\
         \phantom{Soft}+DeFeat &           &     79.57   &     60.78     &     55.84 \\
         \bottomrule
    \end{tabular}~
    \begin{tabular}{@{}cc@{}}
         \toprule
         \multicolumn{2}{c}{VEDAI} \\
         \cmidrule(l){1-2}
              AP10     &      AP25\\
          \midrule
             81.04     &     80.91\\
          \midrule
          \midrule
             74.85     &     74.85\\
             73.23     &     73.19\\
             75.66     &     75.63\\
          \bf75.90     & \bf 75.90\\
          \midrule
             76.71     &     76.69\\
             75.19     &     75.19\\
          \bf78.58     & \bf 78.57\\
             74.32     &     74.32\\
         \bottomrule
    \end{tabular}
    \caption{
    Comparing supervised students trained on a half training set
    (indicated by \protect\pie{180}) with self-trained students using only teacher's
    predictions on the full training set (indicated by \protect\pif{360}).
    The teacher's performance
    is added for reference.
    }
    \label{tab:voc_vedai_half_vs_full}
\end{table}

\begin{table}[t]
    \centering
    \begin{tabular}{@{}lcrrrr@{}}
         \toprule
         Teacher                & \pha{360}  & \pie{180} 58.24 & \pie{ 90}    53.11 &  \pie{ 45} 46.42 \\ %
         \midrule
         \midrule
         Supervised             & \pie{180} &    51.08  &     47.16   &     39.82  \\ %
         \phantom{Hard}+MSE     &           &    53.40  &     48.88   &     41.83  \\ %
         \phantom{Hard}+PDF     &           &    54.18  &\bf  50.13   & \bf 42.81  \\ %
         \phantom{Hard}+DeFeat  &           &\bf 54.30  &     49.54   &     42.19  \\ %
         \midrule
         Self-trained           & \pif{360} &    55.64  &     51.82   &     46.59  \\ %
         \phantom{Soft}+MSE     &           &    55.53  &     51.41   &     46.21  \\ %
         \phantom{Soft}+PDF     &           &\bf 56.47  &\bf  51.93   & \bf 46.59  \\ %
         \phantom{Soft}+DeFeat  &           &    55.84  &     51.47   &     46.02  \\ %
         \bottomrule
    \end{tabular}
    \caption{
    Comparing mAP of supervised students trained on annotated data of various size
    indicated by the column headers (\protect\pie{180} a half, \protect\pie{90} a quarter,
    and \protect\pie{45} an eighth of the original VOC07-12 training set) with those self-trained
    using only the teachers' predictions on the full training set \protect\pif{360}.
    The teachers, whose performances are shown for reference, are trained using
    the same partial subsets indicated in the column headers.
    }
    \label{tab:voc_half_quarter_eighth_full}
\end{table}

\subsubsection*{Reducing supervising data}

Following up from the previous experiment,
we study the student's behavior with different
ratios of seen and unseen data by the teacher. In particular,
we reduce the  available annotations from a half~\pie{180} to
a quarter~\pie{90} and an eighth~\pie{45} of 
the overall training set (Fig.~\ref{fig:teacher_subset}B,C),
thus increasing the uncertainties of the teacher's predictions
used as soft labels for training the student. The results are show in 
Table~\ref{tab:voc_half_quarter_eighth_full}.

Reducing the amount of annotations
results in weaker teachers and, consequently, decreases the general performance.
While the supervised students suffer from the decline of available annotations,
the students self-trained with the teacher's soft labels can maintain its
performance on par with or even surpass its teacher which has 1.6x more 
parameters and was trained in a fully supervised way.

\subsubsection*{Self-training in the absence of teacher training data}

In this section, we study the student's performance in the complete absence of 
teacher's training data. While the teacher training split are
kept the same as in the previous experiments, the student is trained on 
disjoint subsets as shown in Fig.~\ref{fig:disjoint_teacher_student}.
In Table~\ref{tab:voc_disjoint_half2},
we compare the students self-trained on the same second half
subset~\pii{-180} %
using teachers trained with various disjoint subsets
(Fig.\ref{fig:disjoint_teacher_student}A, B1, C1).

\begin{table}[t]
    \centering
    \setlength\extrarowheight{.3ex}
    \begin{tabular}{@{}lcrrr@{}}
         \toprule
         Teacher               & \pha{360}  & \pie{180} 58.24 & \pie{ 90}    53.11 &  \pie{ 45} 46.42 \\ %
         \midrule
         \midrule
         Supervised            & \pie{180}  &    51.08  &     47.16 &     39.82  \\ %
         Self-trained          & \pii{-180} &     52.86 &     49.63 &     45.09 \\ %
         \phantom{Soft}+MSE    &            &     53.83 &     49.72 &     44.85 \\ %
         \phantom{Soft}+PDF    &            & \bf 54.26 & \bf 50.64 & \bf 45.63 \\ %
         \phantom{Soft}+DeFeat &            &     53.58 &     49.90 &     44.54 \\ %
         \bottomrule
    \end{tabular}
    \caption{
    mAP of students self-trained on unseen data by various teachers
    as in the previous experiments (column headers).
    The students are self-trained
    using the same second training half \protect\pii{-180}.
    }
    \label{tab:voc_disjoint_half2}
\end{table}

It is trivial that the weaker the teachers, the less performance
the students and the general performance is lower than that of
the previous experiment as the teachers could not generalize well
with unseen data points. The feature distillations still show
the benefit, especially PDF, bringing the performance closer to
the teacher's level.

\subsubsection*{Increasing student unlabeled data}

In this experiment, we increase the student unlabeled training data
using the complementary subset with the teachers' as shown in 
Fig.~\ref{fig:disjoint_teacher_student}B2, C2. For the sake of
completion, we also include the performance of student trained on
the 3-quarter subset \pii{-270}
while the teacher is trained on the first 1-eighth \pie{45}.

From Table~\ref{tab:voc_disjoint_ext}, there is
a gradual diminution in the student performance along with the
shrinkage of the teacher's training size despite the expansion
of the students' training size, showing the importance of
targets' accuracy and, subsequently, the teacher generalizability.
Interestingly, a small reducing of the teacher training size,
from \pie{90} to \pie{45} for the same student size \pii{-270},
results in a large gap in the student performance, suggesting
a toleration threshold for data deficiency of the teachers.
The use of knowledge distillation also helps
improve the student, even to surpass the teacher.

\subsection{Cross-task knowledge distillation}
\label{subsec:exp_crosstask}

So far, the teacher and student are both trained for the same task
of object detection. In this experiment, the idea of cross-task KD is
studied. In particular, we train a teacher for semantic segmentation and
observe if a student network could be benefited for object detection.

\begin{table}[t]
\centering
\setlength\extrarowheight{.3ex}
\begin{tabular}{@{}lcccc@{}}
  \toprule
  Teacher              & \pie{180}      58.24 & \pie{ 90} 53.11      &  \pie{ 45}  46.42     &\pie{ 45}  46.42     \\ %
  \midrule                                                              
  \midrule                                                              
  Supervised           & \pie{180}      51.08 & \pie{180}      47.16 &  \pie{180}      39.82 &\pie{180}      39.82  \\ %
  Self-trained         & \pii{-180}     52.86 & \pii{-270}     51.23 &  \pii{-270}     46.28 &\pii{-305}     46.23 \\ %
  \phantom{Soft}+MSE   & \pha{-180}     53.83 & \pha{-270}     51.15 &  \pha{-270}     45.98 &\pha{-305}     46.24 \\ %
  \phantom{Soft}+PDF   & \pha{-180} \bf 54.26 & \pha{-270} \bf 51.34 &  \pha{-270} \bf 46.44 &\pha{-305} \bf 46.53 \\ %
  \phantom{Soft}+DeFeat& \pha{-180}     53.58 & \pha{-270}     51.02 &  \pha{-270}     45.94 &\pha{-305}     45.99 \\ %
  \bottomrule
\end{tabular}
\caption{
    mAP of students self-trained on unseen data by various teachers.
    The students' training sets are disjoint with the teachers': student on the
    second half~\protect\pii{-180} and teacher on the first half~\protect\pie{180},
    the last 3-quarter~\protect\pii{-270} and the first quarter~\protect\pie{90},
    the last 7-eighth~\protect\pii{-305} and the first eighth~\protect\pie{45}, etc.
}
\label{tab:voc_disjoint_ext}
\end{table}

The segmentation teacher uses the same 
architecture~\cite{Zhang2021PDF} with the detection teacher's
as in the previous sections and has the detection head
(localization and classification)
replaced by an FPN segmentation head~\cite{Kirillov2019panopticFPN}.
Feature-imitation KD is also done using the neck features as
before. However, since the teacher's segmentation prediction is not
immediately compatible with student's object detection output,
the PDF-Distill method~\cite{Zhang2021PDF} cannot be applied and
only the results for +MSE and +Defeat are
reported. The student detection is trained using hard labels from
its respective training sets.

\begin{table}[t]
    \centering
    \begin{tabular}{@{}lrrrr@{}}
         \toprule
         Teacher                   &&\hrightk\ 68.39 & \hright\ 71.12 & \hleftk\hright\ 72.40 \\ %
         \midrule
         \midrule
         Supervised       & \hleftk &          40.58 &          40.58 &     40.58 \\ %
         \phantom{Hard}+MSE     &&          42.54 &          42.55 &     42.91 \\ %
         \phantom{Hard}+DeFeat  &&     \bf  43.62 &      \bf 44.23 & \bf 44.73 \\ %
         \midrule
         Supervised       &\hleft   &          47.41 &          47.41 &     47.41 \\ %
         \phantom{Hard}+MSE      &&          48.53 &          49.04 &     49.36 \\ %
         \phantom{Hard}+DeFeat   &&     \bf  48.79 &      \bf 49.56 & \bf 49.79 \\ %
         \bottomrule
    \end{tabular}
    \caption{
    Object detection performance with knowledge distillation (KD) 
    from various semantic segmentation teachers.
    The teachers are trained with disjoint semantic sets of different
    sizes~\protect\hrightk\ and~\protect\hright, and a larger set overlapping with
    the student training set~\protect\hleftk\protect\hright\ .
    The teachers' IOU scores are provided for reference.
    Although the teachers had not been trained on the same data nor the same
    task with the students, the results with knowledge distillation are constantly
    better than without by large margins.
    }
    \label{tab:voc_seg_det_new}
\end{table}

We follow the common practice for training semantic segmentation on Pascal VOC
and include the extra semantic annotations provided by SBD~\cite{Hariharan2011SBD}
to the training set. All the training images are randomly sampled into 2 subsets,
one for detection \hleft\ whose semantic annotations are held back or not available
and the other for semantic segmentation \hright\ whose bounding-box annotations
are withheld, resulting in 7,558 and 7,656 respectively.
Half of the detection images are further withheld to simulate the little annotated
data scenario \hleftk.
For validation, the originally provided validation set for semantic segmentation
with both task annotations are used with 1,443 images. 
The six images with only semantic segmentation are excluded from validation.

From Table~\ref{tab:voc_seg_det_new}, the students with knowledge distillation 
perform better than those without. Imitating features of a teacher,
despite from different task topology, and being trained with few or many,
seen or unseen data, show benefit and correlation to the student performances.

\subsection{Multi-task learning for data exploitation}

\begin{table}[t]
    \centering
    \setlength\extrarowheight{.3ex}
    \begin{tabular}{@{}llcccc@{}}
        \toprule
                                           && \multicolumn{2}{c}{Detection}   &\multicolumn{2}{c}{Segmentation} \\
        \midrule
        Teacher                            && \hleftk          &       49.99  &&      -           \\ %
        \midrule
        \midrule
        Supervised                         && \hleftk          &       40.45  & \hright& 65.04    \\ %
        $\quad$ pretrained on \hright      && \hleftk          &       41.00  &        &          \\ %
        \midrule
        Self-trained                       && \hlefte          &       43.08  &        &          \\ %
        $\quad$ pretrained on \hleftk      && \hlefte          &       44.15  &        &          \\ %
        \midrule
        Self-trained                       && \pif{360}        &       45.80  &        &          \\ %
        $\quad$ pretrained on \hleftk      && \pif{360}        &       46.50  &        &          \\ %
        \midrule
        Multitask                          &  \hleftk\hright  &&       45.63  &        &    65.53 \\ %
        $\quad$ pretrained on \hleftk      &  \hleftk\hright  &&       45.58  &        &    65.23 \\ %
        $\quad$ self-training DET          &  \hlefte\hright  &&       43.84  &        &\bf 70.53 \\ %
        $\qquad$ pretrained on \hleftk     &  \hlefte\hright  &&       45.47  &        &    69.83 \\ %
        $\qquad\quad$ + PDF \hleftk        &  \hlefte\hright  && \bf   47.77  &        &    67.37 \\ %
        \bottomrule
    \end{tabular}
    \caption{ Compare and combine self-training with multi-task learning on partially annotated data
    for object detection and semantic segmentation (\ie each image is annotated with only a single task).
    Despite not using any hard ground truths during training, self-training improves performance over
    the small-size supervised training. Extra data from semantic-segmentation subset provide further
    boost by large margins for both settings. Combining both setups and knowledge distillation see
    further improvements, on par with the larger network (teacher).
    The semantic segmentation performances are included for the sake of completeness.
    }
    \label{tab:selftrain_multitask}
\end{table}

Following the discovery in Sec~\ref{subsec:exp_crosstask}, in this experiment
we explore the possibility of using multi-task learning as data exploitation.
In particular, the problem is formulated under the partially annotated data
setting, where each image is annotated for only a single task,~\ie the annotated
image sets of the two tasks are completely disjoint.

The results comparing self-training on available data and multi-task learning
are shown in Table~\ref{tab:selftrain_multitask}.
Self-training \textit{without} provided annotated ground truths
using the whole detection subset~\hlefte\ and all images (detection
and segmentation subsets)~\pif{360}\ are shown, with optionally pre-trained
on the available detection ground truths~\hleftk.

Agreeing with previous
studies, despite not using supervised ground truths,
self-training improves results over small data-size supervised training,
with large margins when the training data are expanded to cover also the
semantic segmentation subset.  Multitask learning gives a great boost to
the performance compared to %
self-training with the extra advantage of having also semantic segmentation prediction.
Surprisingly, the combination of both does not immediately yield better performance.
We speculate that the mismatching due to possible errors in pseudo detection
labels from self-training and ground truth semantic segmentation leads to 
difficulty in learning task interrelationships. Adding PDF-Distil knowledge
distillation which disentangles features using prediction agreement shows
slight improvement of detection but lower segmentation.

\section{Discussions and Conclusions}
\label{sec:conclusion}

The paper performs extensive experiments on knowledge distillation
under the self-training paradigm for object detection and compare
with the partially annotated multi-task setup
where extra data with only semantic segmentation annotations are used.
In the scarcity of annotations,
self-training from an ill-trained teacher on unseen
data points shows constant favors over full supervision.
Knowledge distillation, especially PDF-Distil for object detection,
can further improve the performance even when the teacher is trained
on a different task while multi-task training shows improvement
with large margins, further boosted when combined with feature-imitation
knowledge distillation.

Each setting, however, has its own limitations.
While multi-task learning depends on the relationship between the two tasks
and the available annotations for each member task, self-training is bound
by the teacher's performance. One common problem of both setups is the
domain gap between the target and extra data.
The paper assumes the same domain of the extra data used for both
self-training and multi-task learning,~\ie Pascal VOC data, so that
the teacher in self-training would still perform reasonably with new data
and multi-task networks could learn similar concepts in both tasks.
Depending on the gap between student-teacher domains,
self-training would break down and multi-task learning would have
difficulty in learning the salient interrelationship. This is out
of scope of the paper and could be a potentially study for further data exploitation.

\section{Acknoledgements}

This work was supported by the SAD 2021 ROMMEO project (ID 21007759) and the ANR AI chair OTTOPIA project (ANR-20-CHIA-0030).

{\small
\bibliographystyle{ieee_fullname}
\bibliography{macro,IRISA-ICIP23,IRISA-BMVC23}
}

\end{document}